\documentclass{article}

\PassOptionsToPackage{numbers, compress}{natbib}

     \usepackage[final]{neurips_2020}

\usepackage[utf8]{inputenc} 
\usepackage[T1]{fontenc}    
\usepackage{hyperref}       
\usepackage{url}            
\usepackage{booktabs}       
\usepackage{amsfonts}       
\usepackage{nicefrac}       
\usepackage{microtype}      

\title{Disentangled Planning and Control\\ in Vision Based Robotics via Reward Machines}

\author{
  Alberto Camacho \\
    Robotics at Google\\
  \texttt{albercm@google.com} \\
  \And
  Jacob Varley \\
Robotics at Google\\
\texttt{jakevarley@google.com} \\
  \And
  Andy Zeng \\
Robotics at Google\\
  \texttt{andyzeng@google.com} \\
  \And
  Deepali Jain \\
Robotics at Google\\
  \texttt{jaindeepali@google.com} \\
  \And
  Atil Iscen\\
Robotics at Google\\
  \texttt{atil@google.com} \\
  \And
  Dmitry Kalashnikov \\
Robotics at Google\\
  \texttt{dkalashnikov@google.com} \\
}

\usepackage{subcaption}
\usepackage{amsfonts}
\usepackage{amsmath}
\usepackage{bbm}
\newtheorem{property}{Property}
\newtheorem{theorem}{Theorem}

\usepackage[textsize=scriptsize,textwidth=1cm,disable]{todonotes}
\newcommand{\albertosays}[1]{\todo[inline,color=blue!20!white]{\textbf{Alberto says:} #1}}
\newcommand{\atilsays}[1]{\todo[inline,color=red!20!white]{\textbf{Atil says:} #1}}

\usepackage{xspace}
\newcommand{\DQNAS}{DQN(AS)\xspace}
\newcommand{\DQNRS}{DQN(RS)\xspace}
\newcommand{\DQNRM}{DQRM\xspace}

\renewcommand{\paragraph}[1]{\textbf{#1}}

\begin{document}

\maketitle

\begin{abstract}
%

In this work we augment a Deep Q-Learning agent with a Reward Machine (DQRM) to increase speed of learning vision-based policies for robot tasks,
and overcome some of the limitations of DQN that prevent it from converging to good-quality policies. A reward machine (RM) is a finite state machine that decomposes a task into a discrete planning graph and equips the agent with a reward function to guide it toward task completion. The reward machine can be used for both reward shaping, and informing the policy what abstract state it is currently at.  An abstract state is a high level simplification of the current state, defined in terms of task relevant features. These two supervisory signals of reward shaping and knowledge of current abstract state coming from the reward machine complement each other and can both be used to improve policy performance as demonstrated on several vision based robotic pick and place tasks. Particularly for vision based robotics applications, it is often easier to build a reward machine than to try and get a policy to learn the task without this structure.

\end{abstract}



\section{Introduction}

\emph{Deep} reinforcement learning (RL), and in particular DQN, has proven success across different domains (e.g, playing Atari games from pixels \cite{MnihKSRVBGRFOPB15-dqn}, playing board games \cite{Silver1140}, and doing control from raw pixel observations in robotics applications \cite{kalashnikov2018qt}). 
The main practical advantage of DQN over tabular methods resides in the capability of neural networks to generalize during inference.
Despite apparent success,  
it has been widely recognized that deep RL methods are sample inefficient, 
and they often require heavy training to reach an acceptable quality.
To mitigate for this, it is common in robotics to provide the RL agent with a set of \emph{demonstrations} (that can be used for doing imitation learning and bootstraping DQN), heuristics, and additional rewards to further guide exploration.
We evidence an important limitation of standard DQN, caused by having to solve two related subproblems: 1) learning 
latent state features, and 2) learning a policy that is conditioned on such features. DQN solves the two subproblems at once, while we show value in decoupling them. Intuitively, some of these 
features describe state properties at an abstract level and may be indicative of the current stage of a task. This collection of features defines an abstract state, a notion that can be exploited to decompose long-horizon tasks into simpler subtasks. Not surprisingly, abstract states
are an additional supervision signal that complement 
rewards in RL.
Our main insight 
is in revealing how crucial an explicit notion of abstract states can really be. Even in very simple tasks, DQN failed to learn good-quality policies unless abstract states were provided. This phenomenon occurs even when the information encoded in abstract states was already contained within state observations.

We address the two 
subproblems
mentioned above by taking 
a hybrid approach to reinforcement learning, 
where the agent has access to a set of \emph{feature detectors}.
State features 
enable
the agent to keep track of which abstract state is at. This way,
policies need be conditioned on fewer latent features and are easier to learn.
Then, we 
construct dense reward functions
that facilitate learning sub-policies conditioned on abstract states.
Our approach makes use of \emph{reward machines}, 
a mathematical structure
that provides a principled mechanism to
drive the agent along different abstract states, and
equips them with a dense reward structure that guides exploration toward task completion \cite{tor-etal-icml18}. 
We show how to construct a reward machine from demonstrations,
%
in a way that it
centers Q values around zero---which is beneficial to deep learning methods.

Our experiments evidence the limits of standard DQN,
which was unable to learn good-quality policies even in very simple 
vision-based robotics tasks.
We
also showcase
how reward machines
help
overcome such limitations.
%
%
%
%
Reward machines provide
supervision signals coming from
abstract states
and reward shaping.
%
%
Abstract states (resp., rewards) alone can be effectively exploited by DQN to improve on sample efficiency and policy quality, but it is in the combination of using abstract states and rewards together that DQN manifests a substantially improved performance.

This work concludes that, 
in certain applications, 
augmenting the agent with meaningful abstract state features
can drastically improve policy learning.
Conversely, failing to provide the agent with this additional structure may prevent DQN from learning good policies \emph{even in very simple tasks}.
Abstract state representations can be learned \cite{LiWL06}, but doing so may not even be needed in practice.
In a number of robotic applications it is easy to obtain abstract state representations by hand---e.g., by extracting features from a depth camera, or the pose of the end effector.
%
Our work
evidences the practicality of reward machines in vision-based manipulation tasks, and
opens the door to
future research directions to better design and exploit reward machines.


\begin{figure}[t]
    \centering
    \includegraphics[width=0.9\textwidth]{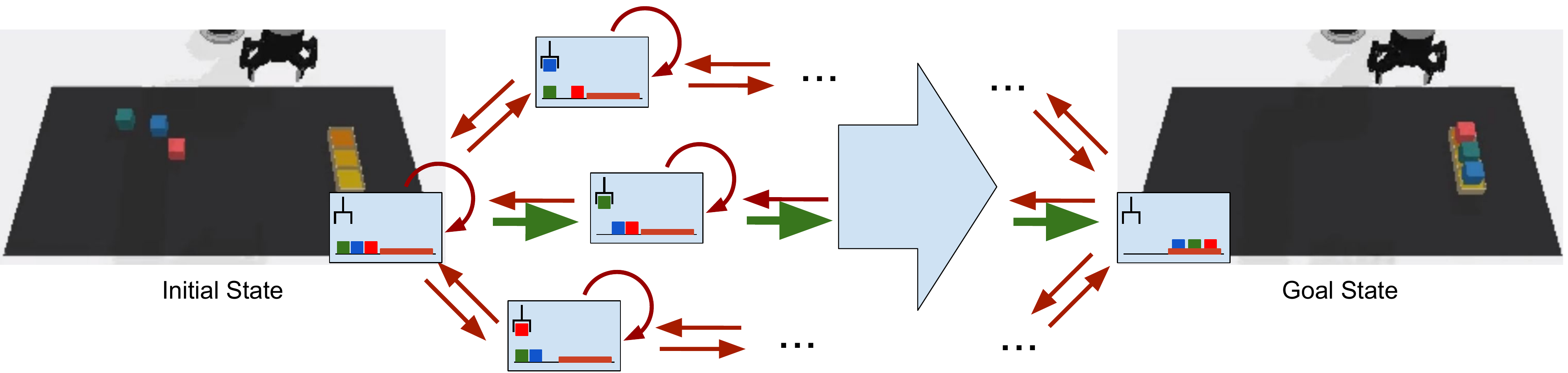}
    \caption{System Overview. Initial and goal states show the starting scene configuration and desired goal configuration.  These map to abstract states within the reward machine (light blue boxes).  In this example, abstract states are determined by the value of $15$ boolean features representing whether or not various blocks are on the table, in different parts of the container, or in the gripper. In our work, the agent is provided a one-hot encoding of the current abstract state (not with the 
    boolean features). Green arrows show the shortest path through the reward machine to reach the goal state. Red arrows illustrate a sample of the many transitions
    between abstract states within the reward machine. 
    } 
    \label{fig:system_overview}
\end{figure}

\paragraph{Running example (kitting task).}
Consider a pick \& place kitting task, in which a robotic arm has to place three blocks of different colors in their designated plate inside of a container (Figure \ref{fig:system_overview}).
Given a robot with depth cameras,
it is feasible to engineer feature detectors that detect whether the end effector is holding a block, the color of such block, and whether a block of certain color is on the table (resp., on their designated plate).
The value of state features can be used to solve this task at an abstract level---e.g., if one block is not on their designated plate, then \emph{pick up} such block; if the end effector is holding a block, then \emph{place} it on their designated plate.
However, state features alone are not sufficient to solve the task, because block locations are relevant to determine (low-level) actions.
Despite the
features alone are not sufficient, their truth values are indicative of task progression.

\section{Related Work}


\textbf{Task Decomposition} 
in RL reduces
complex tasks into simpler subtasks that an agent can learn independently and in parallel---which results in more efficient learning \cite{karlsson1994task}. Hierarchical RL (HRL) is one such framework that 
decomposes planning and control into two alternating levels: 1) a high-level policy that transforms observations into temporally extended actions, and 2) low-level policies (conditioned on high-level actions) that execute for a specified amount of time before control returns to the high level. 
Recent options-based methods \cite{sutton1999between} learn options from reference data \cite{merel2018hierarchical, peng2019mcp} through interactions with 
the environment~\cite{daniel2013learning, co2018self}.
Designing good options still requires substantial manual engineering and hand tuning, and 
is
difficult to scale as the complexity of the task increases.


\textbf{State Abstractions}
are used to reduce the search space in MDPs, by aggregating states that share certain properties in common.
%
A number of techniques have been studied to compute and learn MDP abstractions \cite{LiWL06,AbelHL16}---e.g., bi-simulations \cite{GivanDG03}---,
with trade-offs between state compression and RL performance \cite{AbelAAJLW19}.
%
In this work, we are not concerned with planning in the abstract state. Rather, we
augment state observations with abstract states, and
examine to which extent they are a useful (and sometimes, crucial)
extra supervision signal for deep RL.

\textbf{Reward Machine} (RM) is a framework that decomposes complex tasks into finite-state machines that encode reward functions for MDPs \cite{tor-etal-icml18}. An RM is characterized by a set of internal abstract states that form a graph, in which the agent navigates through different modes of operation. 
Transitions between abstract states indicate different sub-policies, each associated with an individual reward function.
The graph structure of RMs can be exploited toward more sample-efficient RL \cite{tor-etal-icml18,cam-tor-kla-val-mci-ijcai19,tor-etal-neurips19}.
%
The abstract states of an RM do not necessarily need to be directly correlated with the progression of any one specific task.
This property makes them
practical to use for real robot systems, with which various simple hardware signals can be used as features---e.g., end effector states, gripper opening, suction flow sensor readings, proximity sensor readings, joint encoders, etc. 
\albertosays{Revise the parsing of the last sentence.}

\textbf{RL for Vision-Based Manipulation} policies that map pixels to actions have shown promise for of tasks including grasping \cite{kalashnikov2018qt, viereck2017learning, levine2018learning, lenz2015deep, wu2019pixel}, stacking \cite{zeng2018learning}, pushing \cite{popov2017data}, tossing \cite{zeng2020tossingbot}, and even a Rubik's cube \cite{akkaya2019solving}. 
Many of these vision-based RL systems leverage handcrafted features beyond pure rgb observations.
%
For example, the gripper height and status \cite{kalashnikov2018qt}, a hash of the current state \cite{ecoffet2019go}, 
and a CNN prediction of the current pose and state of the cube \cite{akkaya2019solving}. 
Beyond just incorporating additional features into the observation, we demonstrate RMs as a principled and effective means to integrate additional state features to improve the learning efficiency of RL for manipulation tasks.

\section{Background}

A \emph{Markov Decision Process} (MDP) is a tuple $\langle S, s_0, T, A, r, \gamma \rangle$, where $S$ is a set of states, and $s_0 \in S$ is the initial state.
RL agents interact with an MDP by performing \emph{actions}, from the set $A$.
As a result of performing action $a \in A$ in state $s$, the state is updated to some $s'\in S$ according to the probability distribution $T \sim T(s,a)$.
Upon transitioning to $s'$, the agent receives a reward signal $r(s, a, s') \in \mathbb{R}$.
%
RL agents do not know the model, but can \emph{sense} the state to obtain 
an observation $obs(s)$ that may or may not fully describe $s$, and that may or may not be noisy.
The objective of the agent is to learn a policy $\pi$, or a mapping from state observations into action distributions, that maximizes the expected cumulative discount value $V_\pi(s_0) = \mathbb{E}_{\pi} \Sigma \gamma^t r(s_t, a_t, s_{t+1}) \text{~s.t.~} a_t \sim \pi(obs(s_t))$.
The discount factor $0<\gamma\leq 1$ incentivizes the agent to accrue reward early on.
For simplicity, in the sequel we consider fully observable MDPs
where $obs(s) = s$.
%
%
In this paper we address \textbf{behavioral MDPs} in which the objective is to complete a task, and policy execution terminates upon task completion.
Those states for which the task is completed are denoted as \emph{goal} states.
Rewards in behavioral MDPs are designed so that optimizing $V_\pi(s_0)$ is correlated with optimizing for task completion---e.g., rewards can be $1$ in transitions to goal states, and zero otherwise.
%
We presume that the environment is reset after the task is completed, or after the number of steps in the current episode surpasses a fixed threshold.

\paragraph{Deep Q learning for MDPs (DQN).}
Q learning is a model-free RL method that computes a policy $\pi$ by learning a Q function $q_\pi$ over state-action pairs $(s, a)$ that
estimates the expected cumulative reward of performing $a$ in $s$, and acting according to $\pi$ thereafter.
Policy $\pi(s)$ follows a probability distribution over actions induced by $q_{\pi}(s, \cdot)$---e.g., a
\emph{greedy} policy $\pi_\mathit{greedy}(s) = \arg\max_a q(s,a)$.
Q learning interleaves exploration with training. 
Usually, exploration is performed with an epsilon-greedy policy, and experience tuples $(s, a, r, s')$ are stored in an \emph{experience replay} buffer.
Training steps update the incumbent Q function $q$ by sampling batches of experience,
and performing the updates $q(s,a) \leftarrow \alpha (r(s, a, s') + \max_{a'}\gamma q(s', a')) + (1-\alpha) q(s,a)$
for some learning rate $\alpha$.
In behavioral MDPs, we define $ q(s,\cdot) := 0$ in goal states.
%
The greedy policy $\pi_\mathit{greedy}$ is guaranteed to converge to optimal, provided that all state-action pairs are visited infinitely often, and Q values are updated infinitely often.
%
\emph{Deep Q learning} (DQN) extends Q learning with a neural network that acts as Q function approximator \cite{MnihKSRVBGRFOPB15-dqn}. DQN does not have optimality guarantees, but in practice it can scale to larger problems than Q learning thanks to the the inference capabilities of neural networks.

\paragraph{Reward machines.}
%
Formally, a RM is a tuple $\langle U, u_0, \Sigma, \delta, \rho \rangle$, where
$U$ is a set of RM states, and
$u_0$ is the initial state.
A RM operates in an MDP $\langle S, s_0, T, A, r, \gamma \rangle$.
In this paper, $\Sigma = S$ is the set of state observations in the MDP.
The RM issues reward signal in response to MDP transitions $(s, a, s')$, that depends on their current internal state $u$.
More precisely, such reward is $\rho_{u}(s, a, s') \in \mathbb{R}$.
Upon observing a symbol $s' \in \Sigma$, the RM transitions to a new state $u' = \delta(u, s')$.
The mechanics of a RM together with the MDP are equivalent to the mechanics of a 
\emph{cross-product} MDP $M'= \langle  S', s_0', T', A, r', \gamma \rangle$, defined in the natural way. The set of states is $S' = S \times U$, and $s_0' = (s_0, u_0)$.
The transition of state $((s,u), a)$ by action $a$ is a state $(s', u')$, where 
$s' \sim T(s, a)$ 
follows the distribution in the original MDP
and $u' = \delta(u, s')$
is a transition in the RM.
Finally,
the reward function adds up the rewards coming from the original MDP and the RM. More precisely, 
$r'((s,u), a, (s', u')) = r(s, a, s') + \rho_{u}(s, a, s')$.
For more detailed information, we refer to \cite{tor-etal-icml18}.

\paragraph{Q learning for RMs (QRM).}
Whereas the cross-product of an MDP with a RM is a standard MDP that can be handled by standard methods, 
%
the graph structure of RMs can be exploited with dedicated algorithms. In particular,
\emph{Q learning with RMs} (QRM) and \emph{Deep} QRM (DQRM)
showed an improved sample efficiency over Q learning \cite{tor-etal-icml18}.
The benefits of QRM and DQRM where shown, mostly, in MDPs with
non-Markovian rewards, discrete state features, and discrete actions.
%
%
%
Their major trick is that they \emph{broadcast}
experience
to all RM states.
More precisely, 
experience is obtained by interacting with the original MDP, and
each individual experience
$(s, a, s')$ 
is used to generate \emph{multiple} experience tuples in the cross-product MDP: $((s, u), a, (s', u'))$ for each RM state $u$ and RM transition to $u'$.
%
%
It is worth noting that broadcasting experience pays off in RMs where the agent may visit the same state $s$ in different RM states.
This is likely to occur in RMs that encode \emph{non-Markovian rewards}, 
but it is less apparent in MDPs for Markovian tasks.
%
In this work, we show that RMs can be also used for task decomposition in Markovian tasks,
and they have extra benefits beyond broadcasting experience.
Furthermore, we show that RMs can be also effective in vision-based tasks with large action spaces,
which where not evaluated in previous related work.


\section{Deep Q-Learning with Reward Machines (DQRMs) from Demonstrations}
\label{sec:baseline_rm}

In this and the following sections we address the problem of designing a reward function for behavioral MDPs. 
%
The problem can be recast as designing a RM that operates in conjunction with an MDP that has zero constant rewards.
%
%
%
Constructing RMs requires a thought process to
ensure that the graph structure captures task progression, and the reward function provides guidance and induces the desired behavior---similar to the process of designing reward functions, actions, and state spaces for standard MDPs. 
In order to facilitate such endeavor,
recent work introduced principled methods to construct RMs from a high-level logical specification of the intended behavior of the agent \cite{cam-tor-kla-val-mci-ijcai19}.
%
%
%

In this section we present a method to 
construct RMs from demonstrations.
Demonstrations are sequences of states  $s_0, s_1, \ldots, s_n$ that show how a certain task---e.g., stacking a tower of blocks---can be completed, ending in one of the \emph{goal} states.
Demonstrations can be often acquired with a minimal technical knowledge, compared to constructing RMs by hand or designing logical specifications.
%


\subsection{Constructing RMs from Demonstration}

RMs can be constructed from a set of demonstrations in two stages.
%
First, we construct an \emph{abstract planning graph}.
In a second stage, we equip such graph with a reward function to obtain a RM.
%
We introduce the necessary concepts below.
In the following, we presume that the agent has access to a set of \emph{feature} detectors.
Features are associated with properties that can hold in state observations.
Formally,
features are
boolean propositions, and we denote the set of features with AP---which stands for \emph{atomic propositions}.
%
Let
$F: S \rightarrow 2^{AP}$ 
be the function
that maps state observations $s$ into the subset $\sigma \subseteq AP$ of features that hold true in $s$.
%
We say that $\sigma = F(s)$ is the \emph{abstraction} of state $s$ in AP by $F$. 
We further presume that the set of features are rich enough to discriminate between goal and non-goal states---i.e., goal and non-goal states must be mapped into different abstract states.
For convenience, we say that an abstract state is a goal if it is the abstraction of some goal state.

%

\paragraph{Stage 1: Constructing an abstract planning graph.}
The \emph{abstract demonstration} associated to a demonstration  $s_0, s_1, \ldots, s_n$ is a sequence $\sigma_0, \sigma_1, \ldots, \sigma_n$, where $\sigma_i$ is the abstraction of $s_i$ into AP by $F$, for each $i\in \lbrace 1, \ldots, n\rbrace$.
A set of demonstrations defines the \emph{abstract planning graph} $(V, E)$ as follows.
The set of nodes, $V$, is the set of abstract states $\sigma$ that appear in the abstract demonstrations.
The set of directed edges, $E$, connect $\sigma$ with $\sigma'$ if, and only if, $\sigma$ and $\sigma'$ are two consecutive abstract states in some abstract demonstration.
It is worth noting that our construction presumes that the task is Markovian, and goal states do not depend on the past history. 
In the general case,
our methods could be used in conjunction with
existing methods that construct graph-like structures (such finite-word automata and Moore machines) that capture non-Markovian behavior (e.g., \cite{GiantamidisT16,cam-mci-icaps19}).

\paragraph{Stage 2: Constructing a RM.}
The planning graph gives us almost all the ingredients to construct a RM $\langle U, u_0, \Sigma, \delta, \rho \rangle$.
The set of RM states, $U$, is the set of abstract states $2^{AP}$;
the initial state of the RM is the abstraction of the initial state $s_0$ of the MDP, i.e., $u_0 = F(s_0)$; 
the input alphabet, $\Sigma$, is the set of states $S$;
the transition function simply maps observations into abstract states, i.e., $\delta(u, s) = F(s)$;
finally, the reward function $\rho_u(s, a, s') = \mathbbm{1}_\mathit{goal}(s')$ is the \emph{indicator} function that evaluates to 1 if $s'$ is a goal, and zero otherwise.
We have the following property:

\begin{property}
The RM constructed with the method described above assigns reward 1 to MDP executions that achieve the goal, and zero otherwise.
\end{property}

 \paragraph{Running example (cont.)}
 In the kitting task utilized in this work, RM states are abstractions that indicate whether or not the gripper is holding a block of certain color, and whether or not such blocks are placed on their designated plate (resp., on the table).
 Features can be obtained by handcrafting heuristics, or training CNN's to detect either the presence or pose of objects specific to a task.
 Figure \ref{fig:system_overview} shows a planning graph for this task. RM states correspond to the nodes of such graph, depicted with sketches of the features. In our task, all goal states abstract to the same RM state. The reward function $\rho_u(s, a, s') = \mathbbm{1}_\mathit{goal}(s')$ evaluates to $1$ when the observation $s'$ abstracts to a goal state.
%



\section{Inducing Zero-Valued Optimal Q Values With Dense Rewards}

The behavioral MDP
constructed with the RM in Section \ref{sec:baseline_rm}
has very sparse rewards.
In this section we construct RMs with denser rewards that can provide better guidance, via \emph{reward shaping}.
%
%
Reward shaping is a technique that transforms the reward function $r$ in MDPs, with the purpose of obtaining a new MDP that can be better handled by RL and planning methods \cite{NgHR99}.
The new MDP looks 
the same as the original, except for their reward function $r'$.
Reward shaping in infinite-horizon discounted MDPs preserves optimal solutions when reward transformations are
a difference of \emph{potentials}---i.e., 
$r'(s, a, s') = r(s, a, s) + \gamma Pot(s') - Pot(s)$
for some bounded \emph{potential} function $Pot: S \rightarrow \mathbb{R}$.
%
%
%
%
%
%
We use the reward transformation 
presented in
\cite{cam-che-san-mci-socs17,cam-tor-kla-val-mci-ijcai19}, which simply involves modifying the reward function of the RM, $\rho$, with an auxiliary function $Pot: U \rightarrow \mathbb{R}$, that assigns potentials to RM states.
\[
    \rho'_u(s, a, s') = \rho_u(s, a, s') + \gamma Pot(u') - Pot(u), \text{~where~} u' = \delta(u, s')
\]
In addition, we require $Pot(u) = c$ be constant in goal states.
When we apply such reward shaping to the behavioral MDP $M$
constructed in Section \ref{sec:baseline_rm}, we obtain a new behavioral MDP $M'$ with a denser reward function $r'$ that preserves optimal and sub-optimal solutions (Theorem \ref{theorem2}).
This property follows by analyzing the telescopic sum of rewards of policy executions in $M'$---which have finite or infinite length depending on whether the goal is achieved or not, respectively.
The former case simplifies to 
$\Sigma_0^{n-1} \gamma^k r'(s_k, a_k, s_{k+1}') = -Pot(u_0) + \gamma^n c + \gamma^{n-1}$,
and the latter simplifies to
$\Sigma_0^\infty \gamma^k r'(s_k, a_k, s_{k+1}') = -Pot(u_0)$.
Compared to the rewards if executions $M$: (i) rewards to executions that achieve the goal are rescaled from $1$ to $c+1/\gamma$; and (ii) execution rewards are uniformly affected by a constant offset, $ -Pot(u_0)$. Either transformation preserves optimal solutions.



\begin{theorem}
\label{theorem2}
Potential-based reward shaping in discounted behavioral MDPs preserves optimal and near-optimal solutions, provided that the potentials in RM goal states are constant.
\end{theorem}

\paragraph{Optimal Reward Shaping}
We now show how to choose a suitable potential function, $Pot^*(u)$,
when the two properties stated in Theorem \ref{thm:theorem2} (they hold in our running example). We do it in a way that, by construction, the optimal $Q$ values in $M'$ evaluate to zero in all states--i.e., $\max_a q^*(s,a) = 0$ in all $s$.
%
In non-goal states,
we set 
$Pot^*(u) = \gamma^{dist(u)}$, where $dist(u)$ is the minimal distance from $u$ to a goal state in the abstract planning graph, and $\gamma$ is the MDP discount factor. 
In goal states, we set $Pot^*(u) = c$ with $c$ so that $1 + \gamma c = \gamma$.
In practical terms we are forcing that the combination of the rewards coming from $\rho$ and $\rho'$ behaves as if the MDP $M'$ had rewards coming from only a single RM with potentials $Pot^*(u) = \gamma^{dist(u)}$, also in goal states---i.e., $\gamma Pot^*(u) = 1$ in goal states.

The MDP $M'$ with potential function $Pot^*(u) = \gamma^{dist(u)}$ has,
by construction, an optimal value function $V^*(s) := \max_a q^*(s,a) = 0$ (Theorem \ref{thm:theorem2}).
This is because
the immediate rewards for optimal actions in $M'$ are zero ($\gamma Pot^*(u') - Pot^*(u) = 0$).
The same potential function $ Pot^*$ had been used in \cite{cam-tor-kla-val-mci-ijcai19}, 
but the their 
properties on translating optimal Q values to zero where not identified.

\begin{theorem}
\label{thm:theorem2}
Potential-based reward shaping in the MDP $M$ with $Pot(u) = \gamma^{dist(u)}$ as described above results in an MDP $M'$ with zero-valued optimal Q values, provided that:
(i) all feasible RM transitions can be realized with a single action of the MDP; and (ii) all feasible RM transitions can be realized deterministically, by choosing the right action.
\end{theorem}

We observe that doing RL in an MDP whose optimal Q values are zero can be advantageous, especially to \emph{deep} RL approaches.
Bringing target Q values within a range around 0 benefits training very large DQNs, as it reduces the Internal Covariate Shift from gradient descent optimization \cite{ioffe2015batch,ba2016layer}. This leads to more stable training since smooth L1 loss gradients are thereby less likely to excessively perturb convolutional network weights \cite{salimans2016weight}.

\paragraph{Running example (cont.)}
In the kitting task, with planning graph illustrated in Figure \ref{fig:system_overview}, we can assign potentials to RM states according to their distance to a goal state. For example, a state $s$ that has two blocks placed in their designated plate (and the third block is on the table) is two steps away from the goal, and has potential $Pot^*(F(s)) = \gamma^2$. Moving optimally from this state (i.e., picking up the remaining block) results in a reward 0. 
Sub-optimal moves incur in a negative reward.

\section{Experimental Results}
\label{sec:result}

\begin{figure}[t]
    \centering
    \begin{subfigure}{0.48\textwidth}
        \centering
        \includegraphics[height=0.108\textheight]{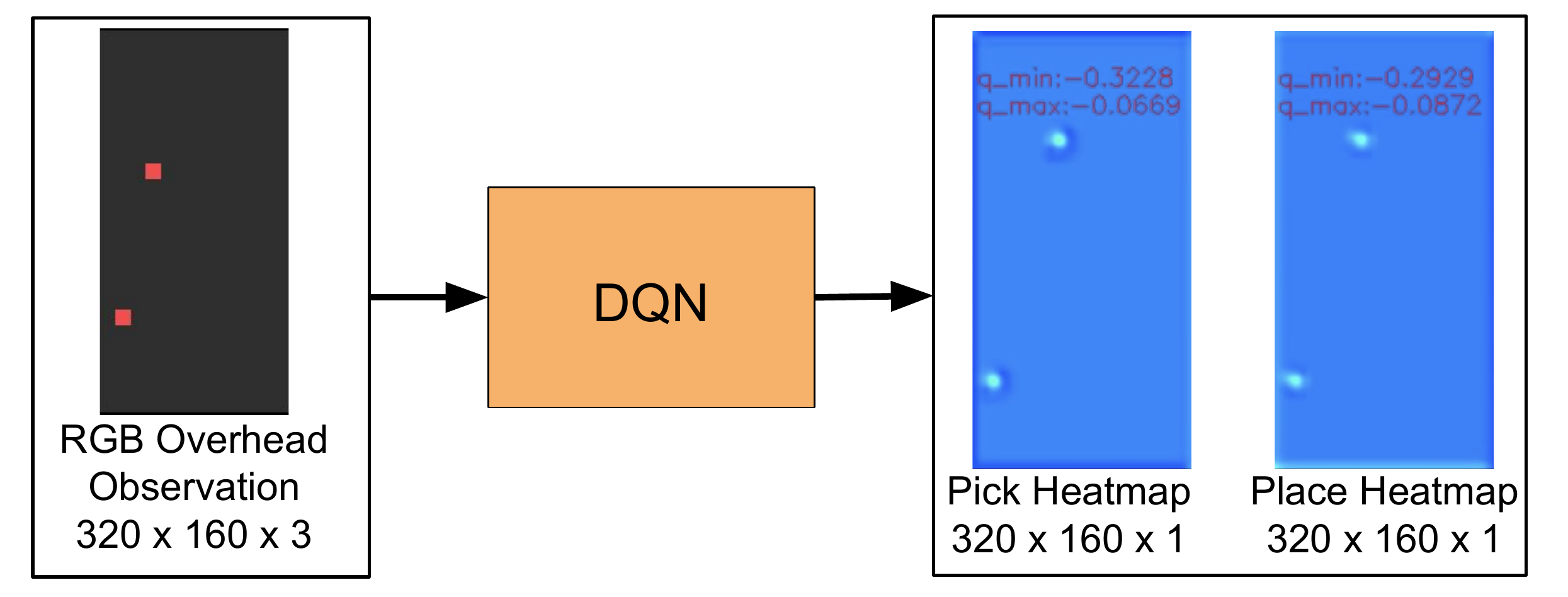}
        \caption{DQN and \DQNRS architectures}
        \label{fig:dqn_architecture}
    \end{subfigure}%
    \hfill
    \begin{subfigure}{0.52\textwidth}
        \centering
        \includegraphics[height=0.108\textheight]{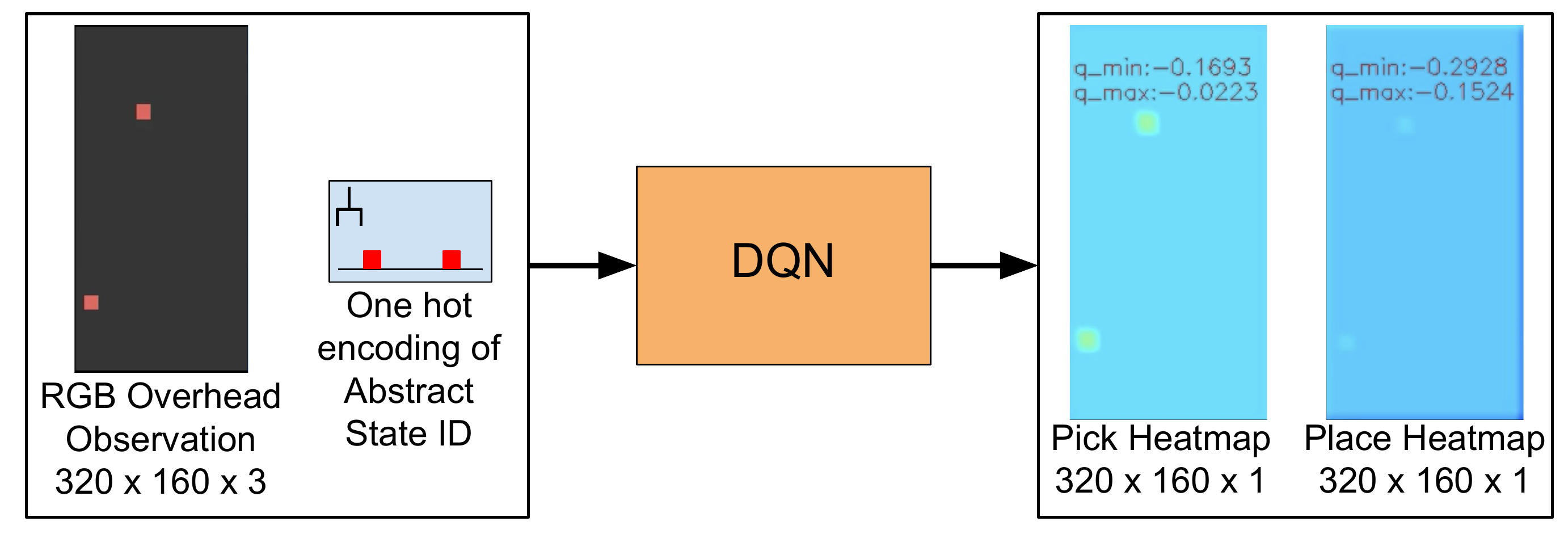}
        \caption{\DQNAS and \DQNRM architectures}
        \label{fig:dqn_rm_architecture}
    \end{subfigure}
    \caption{Overview of different architectures. Figure \ref{fig:dqn_architecture} shows a vanilla DQN architecture, used by DQN and \DQNRS, which takes RGB top down images as observations.
    In DQN and \DQNRS, the RM is used to compute the rewards received by the agent.
    Figure \ref{fig:dqn_rm_architecture} shows the architecture used by \DQNAS and \DQNRM, which take observations augmented with the current RM state.} 
    \label{fig:system_architecture}
\end{figure}

We evaluated the benefits of our methods for task decomposition in two \emph{pick \& place} tasks.
The first task consists on stacking a tower of two red blocks. The second task is the kitting task, with three blocks of different colors, discussed in our running example.
Blocks are initially placed in random locations.
Our environment is simulated in PyBullet, and consists of a \emph{Universal Robot 5} (UR5) grasping arm, a plane, and several blocks of different colors that vary with the task. 
The robot is equipped with RBG and depth cameras with top-down views of the working area. Observations are RGB images with 320x160 pixels per channel.
The robot can perform \emph{pick} and \emph{place} actions, parameterized with the pixel coordinates $(x, y)$ of the target location of the end effector.
The depth camera senses the depth of the objects in the scene, and is used to compute the height $z$ of the target location.
In all our tasks, we constructed reward machines from demonstrations, as explained in previous sections, and with the help of feature detectors that were hardcoded.
%
%
\atilsays{You might want to explain how this setup is realistic (sim-to-real problem?), it's used in other tasks, why it generalizes to other objects, lighting, etc,}

We implemented different versions of DQN in TF-agents \cite{tf-agents}, that
we detail below and illustrate in Figure \ref{fig:system_architecture}. 
In all cases, the DQN consists of two independent feed-forward 43-layer residual networks (ResNets) \cite{he2016deep}.
This architecture has previously been shown to improve learning efficiency for vision-based DQNs \cite{zeng2018learning,hundt2019good,wu2020spatial}.
Q predictions are optimized towards target Q values with Huber loss. Batch training was performed by sampling evenly from an experience replay buffer of size 1000, and a buffer of size 1000 fulfilled with demonstrations. 
%
%
%
Additional details in the Appendix.

\begin{itemize}
    \item DQN: The standard version of DQN, with rewards $1$ in transitions to goal states.
    \item \DQNRS: DQN with the dense reward shaping provided by the RM.
    \item \DQNAS: DQN with observations augmented with the RM state (abstract state).
    \item \DQNRM: The combination of DQN(AS) and \DQNRS.
\end{itemize}

\subsection{Disentangled Planning and Control}

\begin{figure}[t]
    \centering
    \begin{subfigure}[t]{.35\textwidth}
        \centering
        \textsc{Abstract State Classification (Easy)}
        \includegraphics[width=\textwidth]{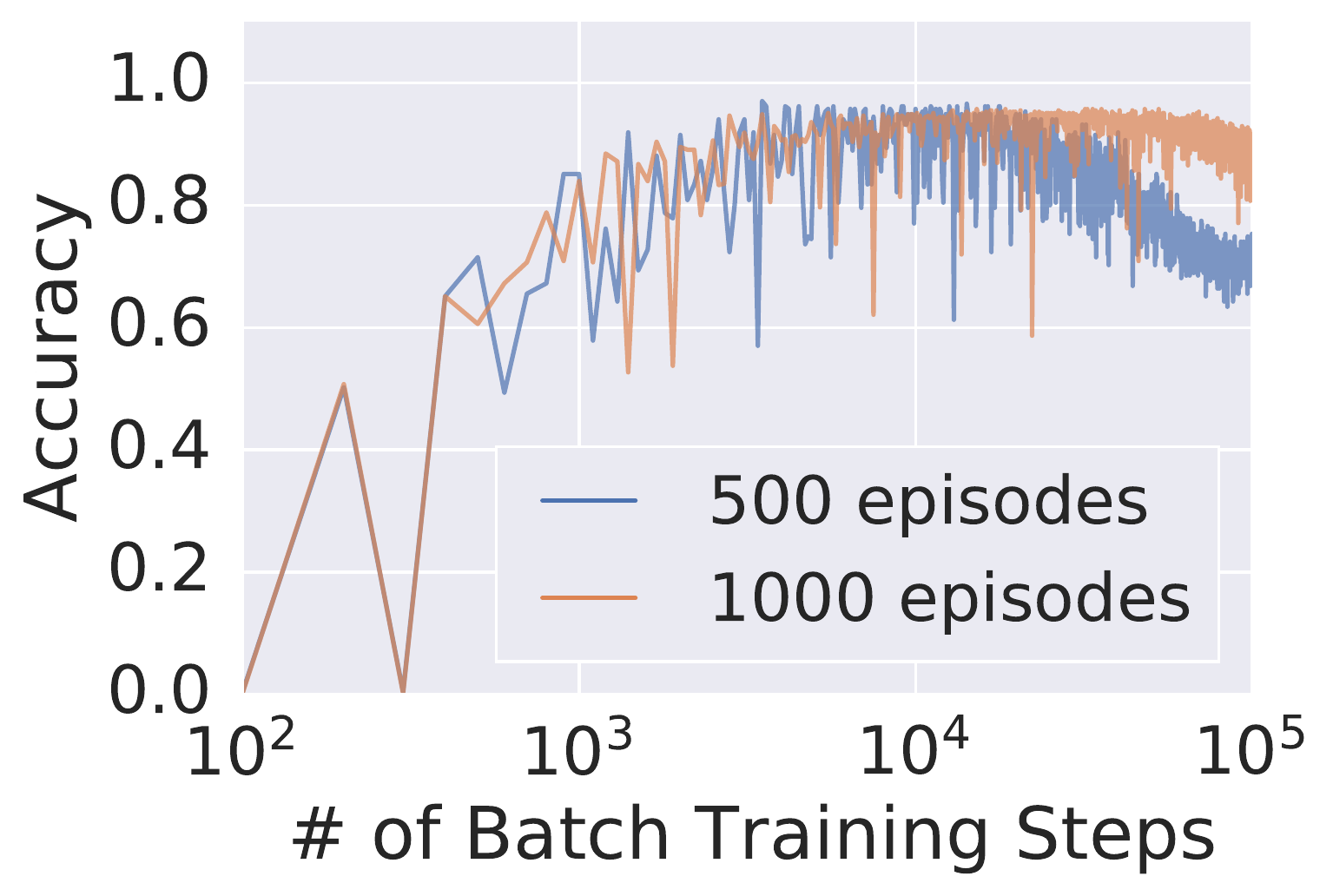}
        \caption{
        ResNet used by DQN has sufficient capacity to classify observations into abstract states with high accuracy, and can be trained easily. Datasets are split for train (80\%) and validation (20\%).
        }
        \label{fig:resnet_classifier}
    \end{subfigure}
    \quad
    \begin{subfigure}[t]{0.28\columnwidth}
        \centering
        \textsc{DQN With Reward Machines (Easy)}
        \begin{subfigure}{\textwidth}
            \centering            \includegraphics[width=0.95\textwidth]{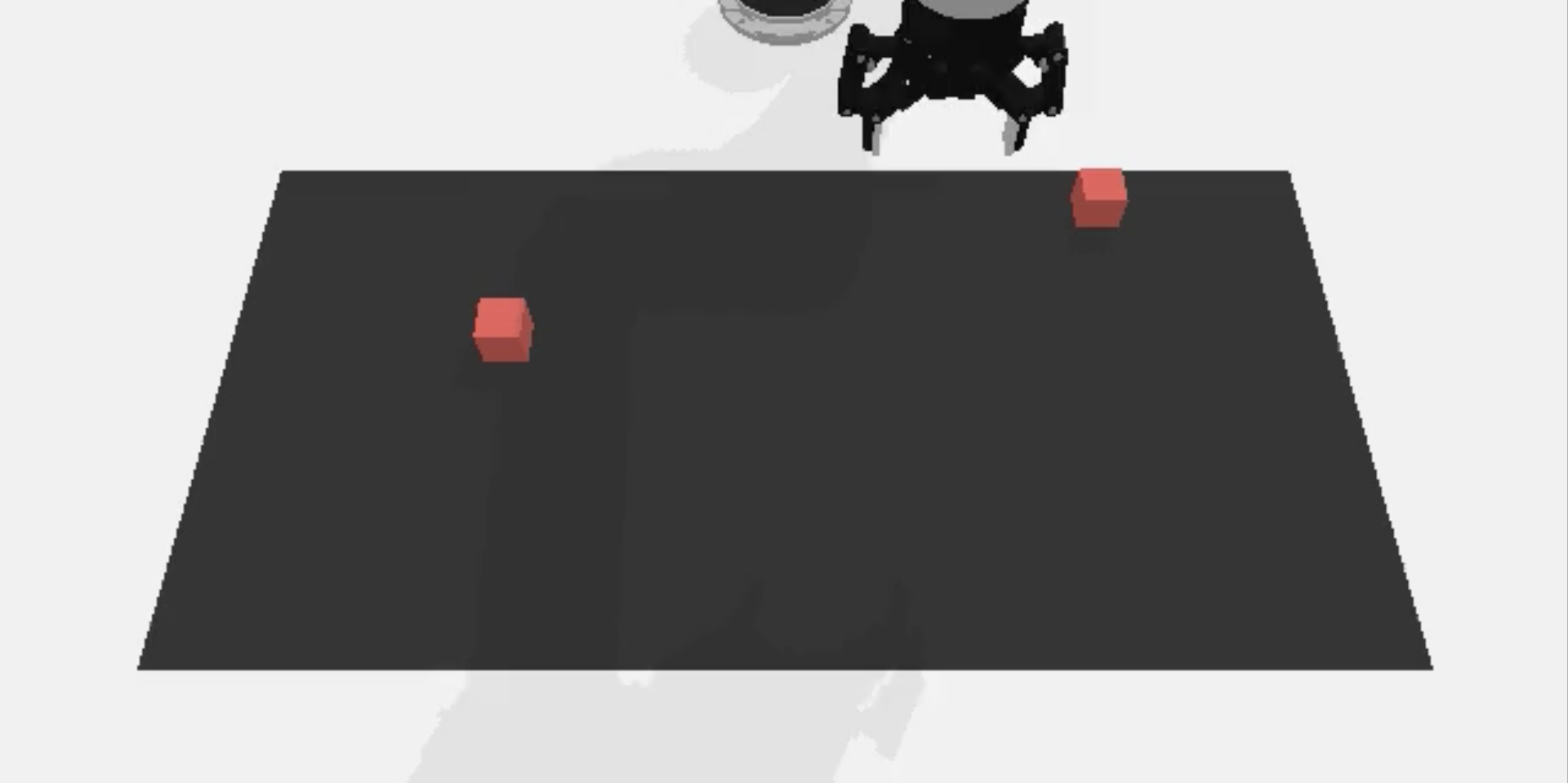}
        \end{subfigure}
        
        \begin{subfigure}{\textwidth}
            \centering
            \begin{subfigure}{0.3\textwidth}
                \includegraphics[width=\textwidth]{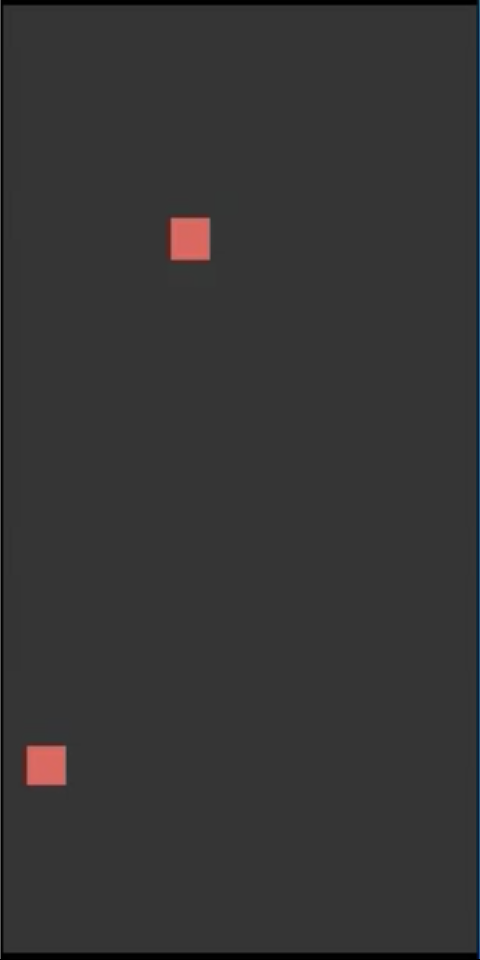}
                \caption*{\tiny RGB \\ Observation}
            \end{subfigure}
            \begin{subfigure}{0.3\textwidth}
                \includegraphics[width=\textwidth]{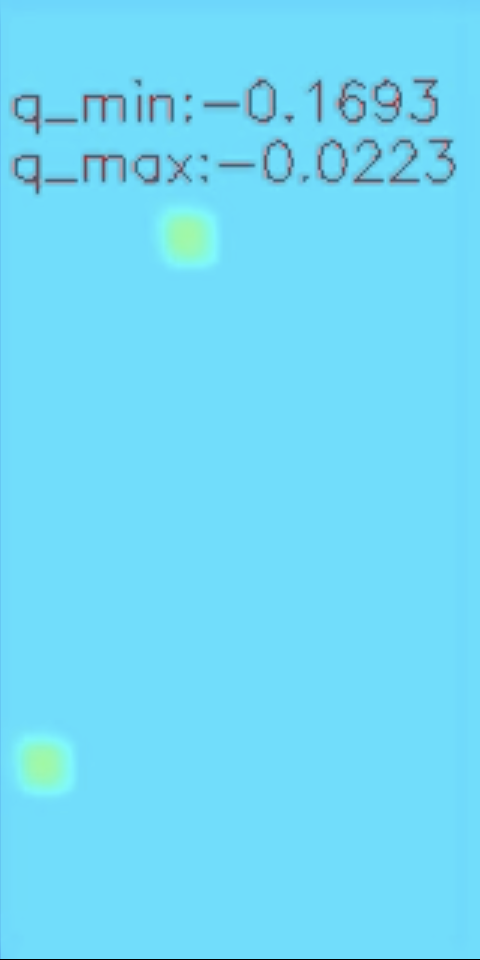}
                \caption*{\tiny Pick Heatmap}
            \end{subfigure}
            \begin{subfigure}{0.3\textwidth}
                \includegraphics[width=\textwidth]{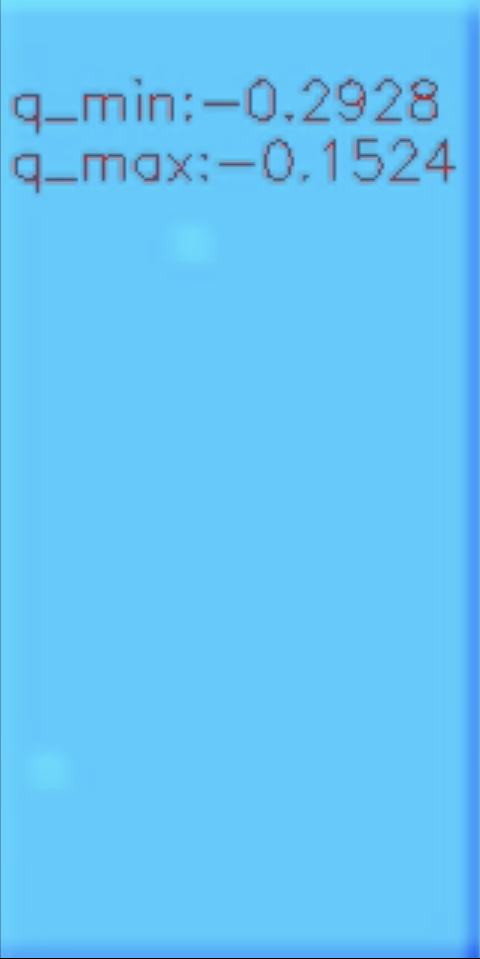}
                \caption*{\tiny Place Heatmap}
            \end{subfigure}
        \end{subfigure}
        \caption{
        DQRM heatmaps
        discriminate high-level actions.
        } 
        \label{fig:DQNAS_pick}
    \end{subfigure}
    \quad
    \begin{subfigure}[t]{0.28\columnwidth}
        \centering
        \textsc{DQN With Reward Shaping (Hard)}
        \begin{subfigure}{\textwidth}
            \centering            \includegraphics[width=0.95\textwidth]{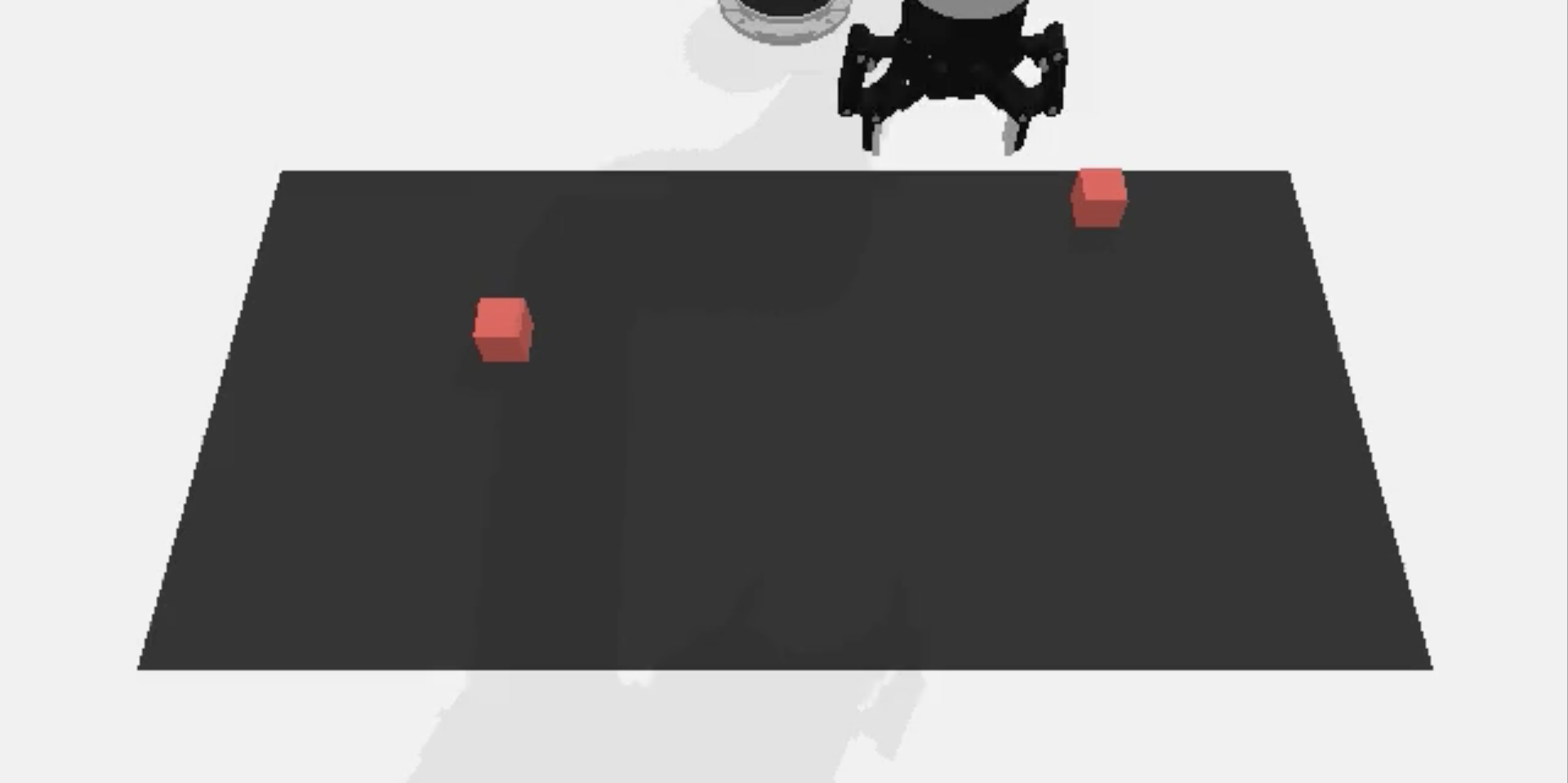}
        \end{subfigure}
        
        \begin{subfigure}{\textwidth}
            \centering
            \begin{subfigure}{0.3\textwidth}
                \includegraphics[width=\textwidth]{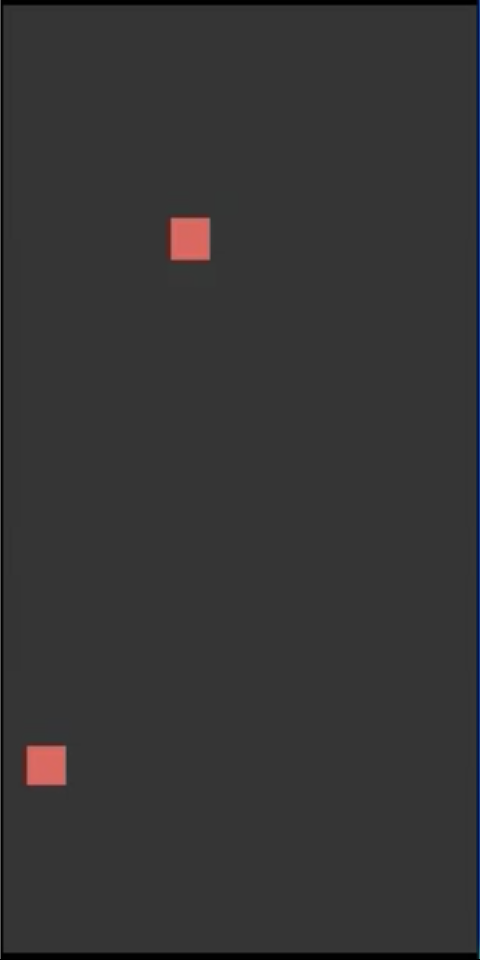}
                \caption*{\tiny RGB \\ Observation}
            \end{subfigure}
            \begin{subfigure}{0.3\textwidth}
                \includegraphics[width=\textwidth]{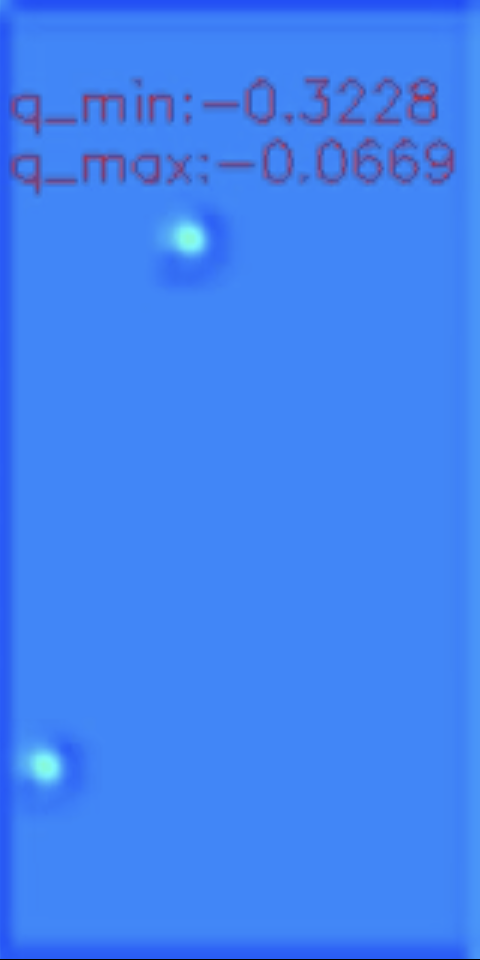}
                \caption*{\tiny Pick Heatmap}
            \end{subfigure}
            \begin{subfigure}{0.3\textwidth}
                \includegraphics[width=\textwidth]{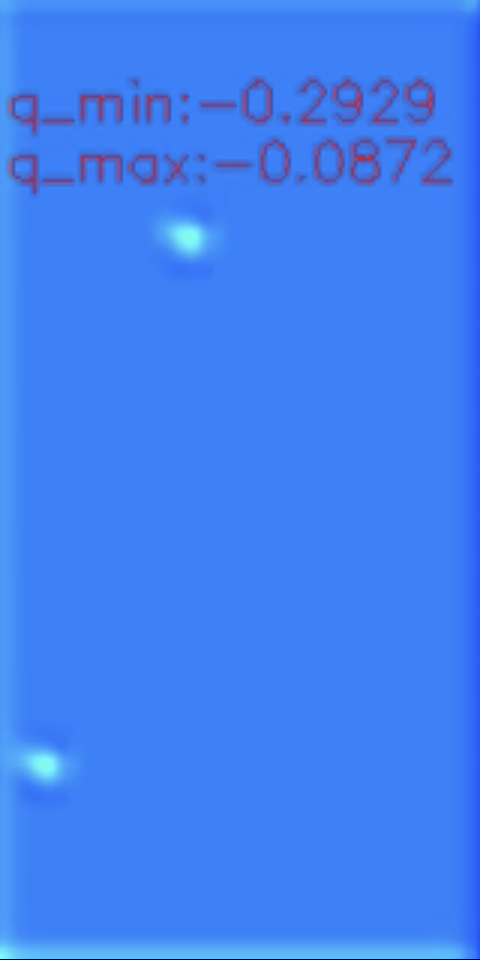}
                \caption*{\tiny Place Heatmap}
            \end{subfigure}
        \end{subfigure}
        \caption{
            \DQNRS Q values do not discriminate high-level actions.
            } 
            \label{fig:dqn_pick}
    \end{subfigure}
    \caption{
    Abstract states are useful state features that help DQN disentangling 
    planning and control.
    DQN can find it challenging to learn a policy that conditions actions on abstract state features, even when learning those features is an easy classification task (Figure \ref{fig:resnet_classifier}).
    \DQNRM takes abstract states as part of the input, and is able to make a clear distinction on the right high-level action \emph{pick}
    (planning)
    and their low-level parameters 
    (control).
    \DQNRS does not take abstract states as part of the input, and cannot discriminate between high-level actions  (Figure \ref{fig:dqn_pick}).
%
    The low performance in \DQNRS is not due to the lack of capacity of the neural network, 
    nor to the difficulty of the classification task.
    }
    \label{fig:what_and_where}
\end{figure}

We evidenced some of the limitations of DQN. When actions are highly conditioned on high-level state features, DQN may not be successful at disambiguating between states that differ at an abstract level. 
Our tests consist of a very simple stacking blocks task: when there are two blocks on the table, the agent should pick up one block; when there is one block on the table, the agent should place the block it's currently holding on top of the other block.
%
%
%
The problem is that when training DQN, the ResNet network cannot distinguish between observations in which there is only one block on the table (the optimal action is \emph{place}) from observations in which there are two blocks on the table (the optimal action is \emph{pick}).
The agent perceives that \emph{sometimes} the good action is \emph{pick}, and \emph{sometimes} is place, and that in all cases the target location of the gripper is a red block.
In Figure \ref{fig:dqn_pick} we see this phenomenon: the Q value heatmaps for the pick and place actions look very similar.
As a sanity check, we assessed that
the low performance of DQN was not due to a limited capacity of the ResNet network. 
Figure \ref{fig:resnet_classifier} evidences that that this is not the case, and the ResNet can be trained to classify observations into abstract states.
Moreover, it appears that the classification task is easy.

Figure \ref{fig:DQNAS_pick} shows that DQRM can successfully disambiguate between abstract states, and output optimal actions. This is because good-quality policies need be conditioned in fewer latent features, given the current abstract state.
It appears that classifying observations into abstract states is easy, and RL given abstract states is also easy, but the combined task of learning policies without abstract states is significantly more difficult.
RMs and DQRM make the combined task easy.

\subsection{Evaluating the Benefits of RMs for Task Guidance}


\begin{figure}[tbh]
    \centering
    \begin{subfigure}{0.95\textwidth}
        \centering
        \includegraphics[height=0.12\textheight]{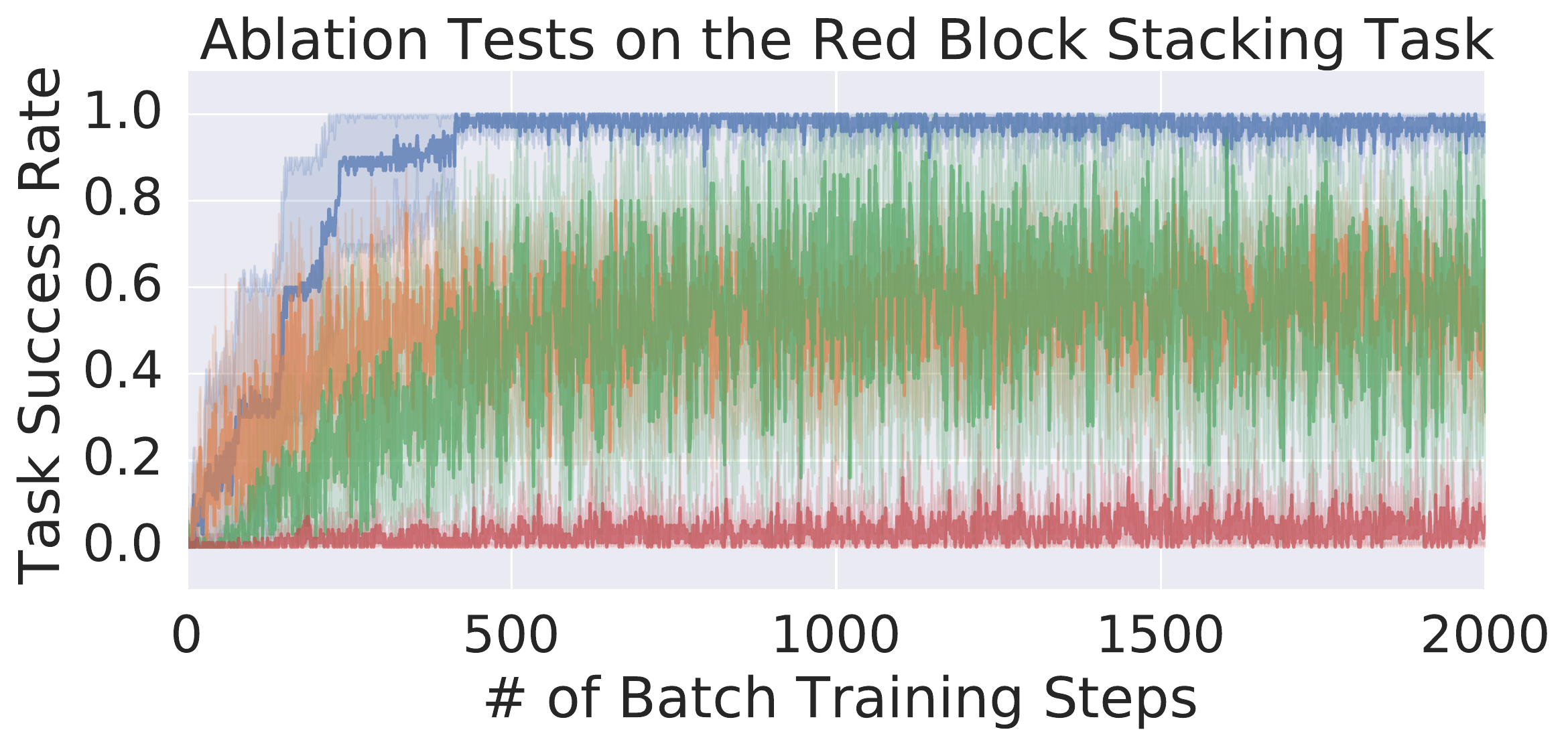}
        \includegraphics[height=0.12\textheight]{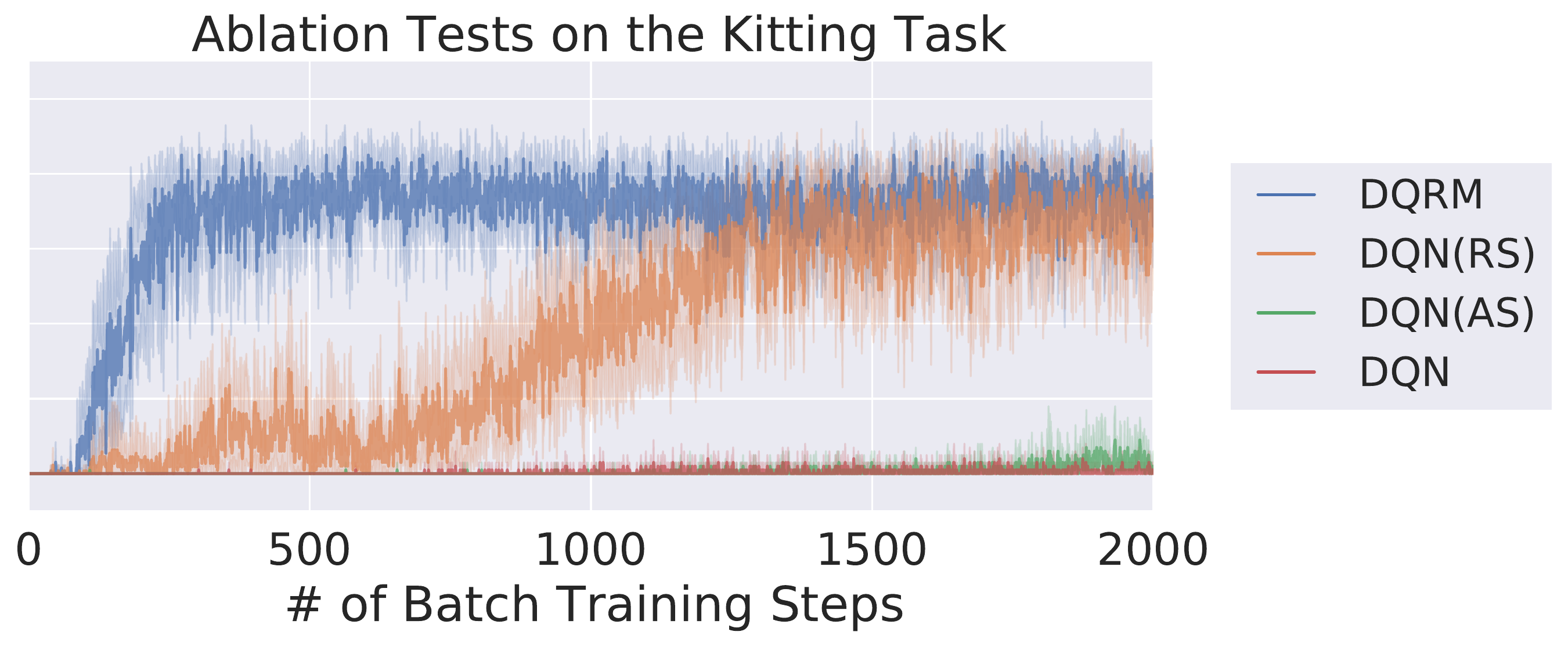}
        \caption{
        Ablation tests of different versions of DQN on two pick \& place tasks.
        The abstract state variables and reward shaping of our \DQNRM approach provide guidance and greatly improve the performance of DQN.
        }
        \label{fig:dqn_ablation}
    \end{subfigure}
    \begin{subfigure}{0.95\textwidth}
        \centering
        \includegraphics[width=0.95\textwidth]{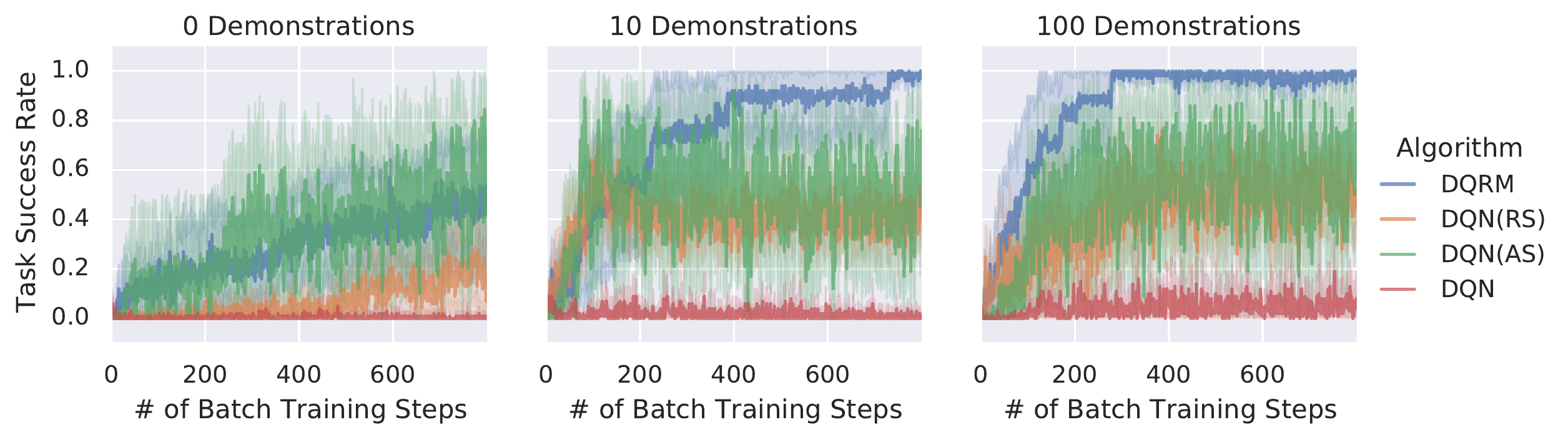}
        \caption{
         Impact of the number of demonstrations (0, 10, and 100) in the performance, evaluated on the block stacking task.
        \DQNRM is the configuration that best exploits the guidance in demonstrations.
        }
        \label{fig:performance_vs_demonstrations_2_red_blocks}
    \end{subfigure}
    \caption{
    Success rate vs. number of episodes for different versions of DQN, trained in batches of size 64 at the end of each episode.
    Results are averaged over 10 experiments, and 10 evaluation runs per datapoint.
    Abstract states, reward shaping, and demonstrations provide useful supervision.
    }
\end{figure}




RMs provide the agent with two sources of guidance (abstract states and reward shaping) that complement each other.
Reward shaping induces a denser reward structure that can improve guidance. 
In our experiments (see Figure \ref{fig:dqn_ablation}), \DQNRS greatly outperformed standard DQN.
Abstract states help correlate optimal actions with observations, 
In the red block stacking task, \DQNAS was successful in doing so thanks to the short makespan of the task (two steps).
However, abstract states alone where not beneficial to \DQNAS in the kitting task (which has a longer makespan of six steps).
A great result in our experiments is that the two sources of guidance mentioned above 
complemented each other. 
We see this in \DQNRM, whose performance greatly outperforms the performance of all other configurations.
While \DQNRS needed 1,500 batch training steps to achieve an acceptable success rate, \DQNRM needed only 200 batch training steps.
Furthermore, \DQNRM could learn better-quality policies with fewer batch training steps, and their success rate is more stable.

\subsection{Exploiting RMs and Demonstrations}


In robotics applications it is common to use a set of expert demonstrations, in conjunction with exploratory data, to help training DQN.
%
%
%
%
\DQNRM can also benefit from demonstrations (see Figure \ref{fig:performance_vs_demonstrations_2_red_blocks}). In our tests on the stacking blocks task, we only needed to provide \DQNRM with a handful of them (10) to see a significant performance increase.
Batch training samples evenly from the demonstration and exploration buffers, and good-quality demonstrations successfully bias learning. 
We only needed about 100 demonstrations to reach a peak of performance in \DQNRM, 
comparable to training with 500 demonstrations (see Figure \ref{fig:dqn_ablation} left).
%
%
Without demonstrations, the configuration that performed best is \DQNAS, with high variance compared to \DQNRM.
However, once some demonstrations are provided to the agent, 
the performance \DQNRM greatly outperformed others.

\section{Conclusion}
\label{sec:conclusion}

This work illustrates the benefits of using reward machines in vision based robotic manipulation tasks with large action spaces, and how they help overcome the limitations of DQN. We analyze two related subproblems that make training DQN policies difficult: learning meaningful latent state features, and learning a policy that is conditioned on such features. Providing the agent with the abstract state helps disentangle planning and control, and the resulting policies are easier to train. Reward machines provide the mathematical structure to manage abstract states in a principled manner, while also providing a dense reward function to give guidance toward task completion. Whereas constructing reward machines by hand can be impractical, we show how to construct them from a set of demonstrations that can also be leveraged to bootstrap DQRM. Either abstract states or dense reward shaping alone can be exploited to improve on sample efficiency and policy quality. Notably, DQRM manifests a substantially improved performance with the combination of using both abstract states and rewards together.  In some cases, DQRM learns good-quality policies with less than 2000 batch training steps, while the sole use of either abstract states or dense reward shaping alone requires 1,500 batch training steps to asymptotically converge to policies that have worse quality.

\bibliographystyle{plain}
\bibliography{main_camacho_deeprl_2020}  
\newpage
\section*{Appendix}

\subsection{Detailed Experimental Setup}
\label{sec:detailed_experimental_setup}

\paragraph{Environment implementation.}
We implemented the testing environments in OpenAI Gym, with the PyBullet physics engine. 
The environment consists of a \emph{Universal Robot 5} (UR5) grasping arm, a plane, and several blocks of different colors that vary with the task. 
The robot is equipped with RBG and depth cameras with top-down views of the working area. Observations are RGB images with 320x160 pixels per channel.
The robot can perform \emph{pick} and \emph{place} actions, parameterized with the pixel coordinates $(x, y)$ of the target location of the end effector.
The depth camera senses the depth of the objects in the scene, and is used to compute the height $z$ of the target location.

\paragraph{Reward machine implementation.}
Reward machines are implemented as an OpenAI Gym class, and are used in conjunction with the environment to compute rewards.
We implemented auxiliary functions to compute reward machines from a set of demonstrations, as explained in the paper.

\paragraph{DQN agents implementation.}
We implemented different versions of DQN in TF-agents \cite{tf-agents}.
In order to use TF-agents with OpenAI Gym environments and reward machines, we made use of wrappers.

\paragraph{Stacking blocks task.}
The stacking blocks task starts with two identical red blocks placed in random locations.
The goal of the task is to stack one block on top of another.

\paragraph{Kitting task.}
The kitting task consists of three blocks of the same size but different colors--red, green, and blue---, and a container with three plates. 
One of the plates situated in one extreme of the container has a slightly different color.
All objects are initially placed at a random location.
The goal of the task is to place each block in their designated container: the red block should be placed inside of the distinct plate; the green block has to be placed in the middle plate; and the blue block has to be placed in the remaining plate.

\paragraph{Demonstrations.}
Demonstrations were handcrafted. They were generated with independent episodes, with objects initially placed in random locations.
In the kitting task, we generated demonstrations that placed blocks in a fixed order: first the red block, then the green block, and finally the blue block.
We found that this fixed order induced slightly better performance in \DQNRM, compared to using demonstrations that placed blocks in arbitrary order.
We conjecture that this improved performance is due to the bias induced by demonstrations in the exploratory policy---which turns to be a good bias.

\subsection{Detailed Model Architecture}
\label{sec:detailed_model_architecture}

\paragraph{ResNet network.}
In all cases, the DQN consists of two independent feed-forward 43-layer residual networks (ResNets) \cite{he2016deep}, each with a fully convolutional encoder-decoder architecture---one that outputs a dense pixel-wise prediction of Q values for picking, and the other for placing (see Figure \ref{fig:system_architecture} in the paper).
Each pixel corresponds to an action (pick or place) executed at that respective location in the scene.
Both ResNets jointly represent the Q function, to which the argmax of the predicted Q value maps returns a pick or place action from a greedy policy.
This architecture has previously been shown to improve learning efficiency for vision-based DQNs \cite{zeng2018learning,hundt2019good,wu2020spatial}, since Q value predictions are thereby spatially anchored on visual features---inducing translational equivariance
(such that if an object to be picked in the scene is translated, then the picking action-values also translate).

\paragraph{Input observations.}
The four DQN agent configurations that we tested (namely, DQN. \DQNAS, \DQNRS, and \DQNRM) are fed with different input observations.
\begin{itemize}
    \item In DQN, the input is a 3-channel RGB image with the top-down view of the scene. In the kitting task, the input to DQN has an additional 1-channel layer with a constant number tied with (but different from) the color of the block being held (or 255, if no block is held).
    \item In \DQNRS, the input is the same as in DQN.
    \item In \DQNAS, the input is the same as in DQN, plus $N$ additional channels (as many as abstract states in the RM). The values of each channel are zero, except for the values of the channel tied with the current RM state, that are set to one---in other words, these $N$ channels are one-hot channels that encode the current RM state.
    \item In \DQNRM, the input is the same as in \DQNAS.
\end{itemize}

\paragraph{Rewards.}
The reward given to each of the DQN configurations that we tested are different, depending on whether reward shaping is performed. In particular:
\begin{itemize}
    \item In DQN, rewards are $+0.5$ if the MDP state transitions to a goal state, and zero otherwise.
    We set rewards to $+0.5$ and not to $+1$ because the range of the values that the ResNet can output is $[-0.5, +0.5]$.
    \item In \DQNRS, rewards are reshaped. Rewards to transitions $(s, a, s')$ are the difference of potentials $\gamma Pot(s') - Pot(s)$. The potentials function $Pot(s) := \gamma^{dist(s)}$ is an exponential decay with the distance from $s$ to the goal in the abstract planning graph.
    In practical terms, we set $Pot(s) = 1$ in goal states,
    as we suggested in the paper.
    \item In \DQNAS, rewards are the same as in DQN.
    \item In \DQNRM, rewards are th same as in \DQNRS.
\end{itemize}

\paragraph{Output actions.}
The agent outputs actions composed of the target $(x, y, z)$ location of the gripper, together with a bit of information telling whether the action is \emph{pick}, or \emph{place}.
Actions are computed from the Q value heatmaps of the ResNet networks.
Each agent has two independent ResNet networks that output, respectively, heatmaps for the pick and place high-level planning actions.
The planning action (pick or place) is selected from the heatmap that has the highest Q value.
The control action---i.e., the low-level coordinates $(x, y, z)$ with the target location of the gripper---are computed from the argmax of the heatmap ($(x, y)$) and the $z$-coordinate of the object placed in the $(x,y)$ location of the scene, obtained from the depth camera.

\paragraph{Training.}
Q predictions are optimized toward target Q values with Huber loss, 
trained with Adam and a learning rate of $10^{-5}$, with a discount factor $\gamma = 0.7$. 
Training steps were performed each 8 exploration steps, with a batch size of 64.
Unless otherwise noted, 
batch training was performed by sampling evenly from an experience replay buffer of size 1000, and a buffer of size 1000 fulfilled with demonstrations.

\paragraph{Data collection.}
We used an epsilon-greedy exploration policy, with 
$\epsilon=0.3$.
The exploration (and policy evaluation) horizon is 8 steps, after which the environment is reset.

\paragraph{Experience replay buffers.}
Each agent has two independent experience replay buffers.
The first one is filled with demonstrations. The second one is filled with exploration data.


\end{document}